# Hardware-efficient Residual Networks for FPGAs

Olivia Weng, Alireza Khodamoradi, and Ryan Kastner
Dept. of Computer Science and Engineering, UC San Diego
{oweng, akhodamo, kastner}@ucsd.edu

*Abstract*—Residual networks (ResNets) employ skip connections in their networks—reusing activations from previous layers—to improve training convergence, but these skip connections create challenges for hardware implementations of ResNets. The hardware must either wait for skip connections to be processed before processing more incoming data or buffer them elsewhere. Without skip connections, ResNets would be more hardware-efficient. Thus, we present the *teacher-student learning method* to gradually prune away all of a ResNet's skip connections, constructing a network we call NonResNet. We show that when implemented for FPGAs, NonResNet decreases ResNet's BRAM utilization by 9% and LUT utilization by 3% and increases throughput by 5%.

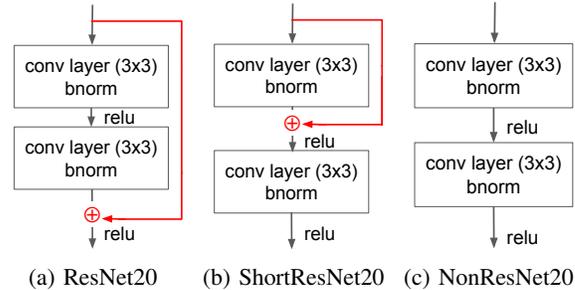

Fig. 1: Residual block (ResNet layers involved in a single skip connection) topologies from our ResNets. ResNets are essentially a series of residual blocks. Skip connection in red.

## I. INTRODUCTION

RESIDUAL networks (ResNets) [3] like ResNet20 are characterized by their skip connections, wherein a layer adds activations from a previous layer to its output (Figure 1a), in order to circumvent the vanishing/exploding gradient problem [1], [2] that occurs when training deep networks. But skip connections create a major issue in hardware implementations because they introduce a bottleneck during inference: the layers whose activations are reused in future layers must wait for this reuse to occur before accepting new input and continuing to compute. One solution is to buffer these activations elsewhere, allowing these layers to begin processing inputs as they are received; however, this increases the model's resource utilization.

To alleviate this inference bottleneck without increasing resource usage, we ideally would want to run a version of a ResNet with its skip connections removed. But removing skip connections either before or after training the network causes its inference accuracy to disintegrate. ResNets cannot handle such drastic changes to its architecture. Thus, we want to modify the architecture of a ResNet gradually to temper the shock of taking away skip connections.

In this paper, we show how to gently build and train a ResNet without any skip connections so that it can be efficiently implemented in hardware without allocating extra buffers or creating a bottleneck during inference. This work focuses specifically on ResNet20 (a ResNet with 20 layers) trained on the CIFAR-10, CIFAR-100, and SVHN datasets and implemented for FPGAs.

To progressively remove a ResNet's skip connections, we introduce the *teacher-student learning method*, a subset of the knowledge distillation learning method [4]. The teacher-student learning method involves having NonResNet20's residual connection layers (Figure 1c) act as "students" that learn from their corresponding ResNet20 residual layers (Figure 1a), who are the "teachers," to achieve a comparable accuracy. To ease the transition from having skip connections to having no skip connections at all, we modify and retrain a ResNet to have shorter skip connections and use this model we call ShortResNet (Figure 1b) as the teacher for NonResNet.

To evaluate the impact of ResNet skip connection removal in hardware, we implement a ResNet20 residual block (Figure 1a) and a NonResNet20 "non-residual" block (Figure 1c) for FPGAs using the FINN library [5] in Vivado HLS.

Our contributions are: (1) introducing the idea of removing skip connections from neural networks for more efficient FPGA neural network implementations, (2) the teacher-student learning method of gradually pruning away skip connections from ResNets, and (3) demonstrating that NonResNets are implemented more efficiently than ResNets in FPGAs, decreasing BRAM utilization by 9% and LUT utilization by 3% and increasing throughput by 5%.

## II. TRAINING NONRESNET20

As previously discussed, one idea is to remove ResNet20's skip connections altogether and see if we can get a comparable accuracy, i.e., train NonResNet20 from scratch using randomly initialized weights; however, with this training method, NonResNet20 is unable to learn an inference accuracy better than randomly choosing a class—we cannot simply remove the skip connections. As a result, we introduce the *teacher-student learning method*, wherein we use ResNet20's learned weights to teach NonResNet20. When learning directly from ResNet20 however, NonResNet20 is still unable to perform better than random classification, so we turn to the idea of

TABLE I: 32-bit Accuracy for various image classification datasets

| Dataset | Accuracy (%) | | |
|---|---|---|---|
| | ResNet20 | ShortResNet20 | NonResNet20 |
| CIFAR-10 | **94.03** | 93.29 (0.74 loss) | 93.61 (0.42 loss) |
| CIFAR-100 | 70.36 | 69.70 (0.66 loss) | **71.63** (1.0 gain) |
| SVHN | 96.57 | 96.67 (0.10 gain) | **96.86** (0.29 gain) |

*shortening* the skip connections first, as a way of gently forming ResNet20's model topology to that of NonResNet20. This shortened skip connection ResNet is ShortResNet20 (Figure 1b). ShortResNet20 performs well, with negligible loss in accuracy when compared with ResNet20. As seen in Table I, ShortResNet20 incurs a minimal accuracy loss of at most 0.74% and, in fact, outperforms SVHN by 0.10% in accuracy. Thus, we build NonResNet20 using ShortResNet20 because of their more similar model architectures.

Using ShortResNet20 as the "teacher" in our *teacher-student learning method*, we build NonResNet20 piece by piece, as detailed in Algorithm 1. To iteratively remove all 9 skip connections from ShortResNet20, we first build an intermediary MODEL, which is ShortResNet20 with its first 3 skip connections removed. Then the ShortResNet20 weights are loaded into MODEL, and we train MODEL. With the trained MODEL, we save its weights and use them in the next iteration when we remove the next three skip connections. This process repeats until all skip connections have been removed, and we are left with a complete NonResNet20 model.

---

**Algorithm 1:** TEACHER-STUDENT LEARNING

1 train SHORTRESNET20 with randomized weights
2 let $W$ be weights of SHORTRESNET20
3 **for** $i$ in 1 to 3 **do**
4      let MODEL be SHORTRESNET20 without skip connections 1 to $i*3$
5      initialize MODEL weights with $W$
6      train MODEL
7      $W \leftarrow$ weights of MODEL
8 **end**
9 let NONRESNET20 be MODEL

---

As seen in Table I, NonResNet20 has minimal accuracy loss compared with ResNet20, incurring only a 0.42% loss on CIFAR-10. In fact, NonResNet20 achieves higher accuracies on CIFAR-100 and SVHN, with gains of 1.0% and 0.29%, respectively. Not only is NonResNet20 outperforming ResNet20 with respect to accuracy, it is now primed for hardware efficiency because it does not need the resources required for any skip connections at all.

## III. SYNTHESIZING RESNET20 AND NONRESNET20 IN FINN

We use Xilinx's FINN [5] HLS library to implement ResNet20 and NonResNet20 in Vivado HLS. At the time of writing this report, we have preliminary implementations of a ResNet20 residual block (Figure 1a) called `resblock` and a NonResNet20 non-residual block (Figure 1c) called `nonresblock`. These implementations for now use FINN's binary neural network library, so they assume 1-bit weights and 2-bit activations. As such, our residual blocks are approximations of what the final 32-bit fixed point HLS versions of ResNet20 and NonResNet20 will be. Since ResNets are a series of similarly sized residual blocks connected together, we continue with the synthesis results of these stand-alone residual blocks.

## IV. RESULTS

As seen in Table II, we find that `nonresblock`'s throughput of 1.415 kHz is a 5% increase over `resblock`'s. `nonresblock`'s resource utilization is across the board lower than `resblock`'s, as expected. Take note that without having to implement or compute a skip connection, `nonresblock` uses 9% fewer BRAMs than `resblock`, decreasing total BRAM utilization from 2.83% to 2.68%, and 3% fewer LUTs than `resblock`, decreasing total LUT utilization from 2.97% to 2.91%.

TABLE II: Synthesis results

| Design | Throughput (kHz) | Percent utilization (%) | | |
|---|---|---|---|---|
| | | FF | LUT | BRAM |
| `resblock` | 1.352 | 0.37 | 2.97 | 2.83 |
| `nonresblock` | 1.415 | 0.35 | 2.91 | 2.68 |

## V. CONCLUSION

Making ResNets like ResNet20 more hardware-efficient for inference on FPGAs by removing skip connections is a worthwhile task. The throughput gains and resource reductions are further amplified by the increase in accuracy over ResNet20 that we find with NonResNet20 on certain datasets, such as CIFAR-100 and SVHN. NonResNet is indeed a topology that lends itself to a better, more efficient FPGA implementation. Looking ahead, we seek to quantize our various ResNet20 models to perhaps 16-bit fixed point and then synthesize the complete 16-bit fixed point versions of them in FINN for further comparison and evaluation.